\documentclass{article}
\usepackage{spconf,amsmath,graphicx}

\usepackage{algorithm}
\usepackage{algorithmic}
\newcommand{\xvec}{{\bf x}}

\usepackage{booktabs}
\usepackage{amsfonts,amssymb}
\usepackage{multirow}
\usepackage{subfigure}
\usepackage{amsthm}
\usepackage{url}            % simple URL typesetting
\usepackage{amsfonts}       % blackboard math symbols

\title{Regularized Conditional Alignment for Multi-Domain Text Classification}
\name{Juntao Hu \qquad Yuan Wu$^*$\thanks{*Corresponding author.}}
\address{School of Artificial Intelligence, Jilin University, Changchun, China}
\begin{document}
%\ninept
%
\maketitle
\begin{abstract}

The most successful multi-domain text classification (MDTC) approaches employ the shared-private paradigm to facilitate the enhancement of domain-invariant features through domain-specific attributes. Additionally, they employ adversarial training to align marginal feature distributions. Nevertheless, these methodologies encounter two primary challenges: (1) Neglecting class-aware information during adversarial alignment poses a risk of misalignment; (2) The limited availability of labeled data across multiple domains fails to ensure adequate discriminative capacity for the model. To tackle these issues, we propose a method called Regularized Conditional Alignment (RCA) to align the joint distributions of domains and classes, thus matching features within the same category and amplifying the discriminative qualities of acquired features. Moreover, we employ entropy minimization and virtual adversarial training to constrain the uncertainty of predictions pertaining to unlabeled data and enhance the model's robustness. Empirical results on two benchmark datasets demonstrate that our RCA approach outperforms state-of-the-art MDTC techniques.

\end{abstract}
\section{Introduction}
\label{sec:intro}

With the emergence of deep neural networks (DNNs), researchers have achieved significant breakthroughs across various applications in machine learning, encompassing computer vision, speech recognition, and natural language processing (NLP). Among these domains, NLP stands out as particularly formidable, primarily due to the inherently discrete nature of language data. Text classification, a foundational task in NLP, boasts widespread applications, ranging from spam detection \cite{ngai2011application} to data mining \cite{aggarwal2012survey}. Over the past decades, DNN-based models for text classification have amassed remarkable accolades \cite{joulin2016bag}. However, it's worth noting that these commendable achievements predominantly hinge upon access to copious volumes of annotated data. In practical scenarios, labeled data may indeed be available across multiple domains, yet the quantities often prove insufficient to adequately train a proficient classifier for one or more of these domains. Moreover, certain domains, such as reviews pertaining to medical equipment, may suffer from a complete absence of labeled data. Additionally, a model trained on one domain consistently exhibits subpar performance when applied to another domain characterized by a distinct data distribution. Hence, it becomes imperative to explore strategies for enhancing classification accuracy within the target domain by harnessing the available resources derived from related domains.

Multi-domain text classification (MDTC) endeavors to tackle the above problem. Contemporary MDTC techniques predominantly rely upon adversarial training and the shared-private paradigm to yield cutting-edge performance \cite{wu2021conditional, wu2021mixup}. More specifically, the adversarial training was initially rooted in image generation \cite{goodfellow2014generative}. Subsequently, its purview expanded to encompass cross-domain alignment, with the objective of capturing domain-invariant features that inherently exhibit both transferability and discriminability \cite{ganin2016domain}. The shared-private paradigm serves as a conduit for learning not only domain-invariant but also domain-specific features \cite{liu2017adversarial}. The domain-specific features prove effective in enhancing the discriminability of the domain-invariant features \cite{bousmalis2016domain}. Nevertheless, these MDTC methodologies remain mired in two primary challenges. Firstly, the pursuit of aligning marginal features entails the lurking peril of misalignment. For instance, in the alignment of two domains, such as car and camera reviews, while the car domain may harmonize seamlessly with the camera domain, a favorable car review may paradoxically correspond with an unfavorable camera review. Secondly, the constraint of accessing merely finite quantities of annotated data during training renders the attainment of an optimal classifier or a joint discriminator endowed with sufficient discriminative capacity unattainable. This constraint also predisposes the model to the perils of overfitting.

To tackle the aforementioned challenges, we propose a Regularized Conditional Alignment (RCA) approach. The RCA method orchestrates the alignment of joint distributions encompassing domains and classes, thereby ensuring that samples with different sentiments will not be aligned to the same feature point in the latent space. Furthermore, we integrate principles derived from semi-supervised learning (SSL) to augment the model's discriminability and robustness. Specifically, we incorporate entropy minimization \cite{grandvalet2005semi} and virtual adversarial training (VAT) \cite{miyato2018virtual}. These techniques serve to relocate decision boundaries to the sparser regions of the latent space and impose the essential Lipschitz constraint upon the model. To gauge the efficacy of our proposed RCA approach, we undertake empirical experiments on two MDTC benchmarks: the Amazon review dataset and the FDU-MTL dataset. The empirical findings conclusively demonstrate the superiority of our method over state-of-the-art approaches.

\section{The Approach}
\label{sec:Appro}

In MDTC, we are given $M$ domains $\{D_i\}_{i=1}^M$, each domain consists of two parts: a limit amount of labeled data $\mathbb{L}_i=\{(\xvec_j^i,y_j^i)\}_{j=1}^{l_i}$ and a large amount of unlabeled data $\mathbb{U}_i=\{\xvec_j^i\}_{j=1}^{u_i}$, where $l_i$ and $u_i$ are numbers of labeled and unlabeled samples, respectively. The $M$ domains are defined on $X\times Y$ where $X$ is the input space and $Y=\{1,...,K\}$ is the label space. In this paper, we mainly deal with the sentiment classification problem such that $K=2$. The task is to learn a model $f:X\mapsto Y$. The objective of MDTC is to improve the performance of the model, measured in this paper as the average classification accuracy among the $M$ domains.

%%%%%%%%%%%%%%%%%%%%%%%%

\begin{figure}
    \centering
    \includegraphics[width=0.8\columnwidth]{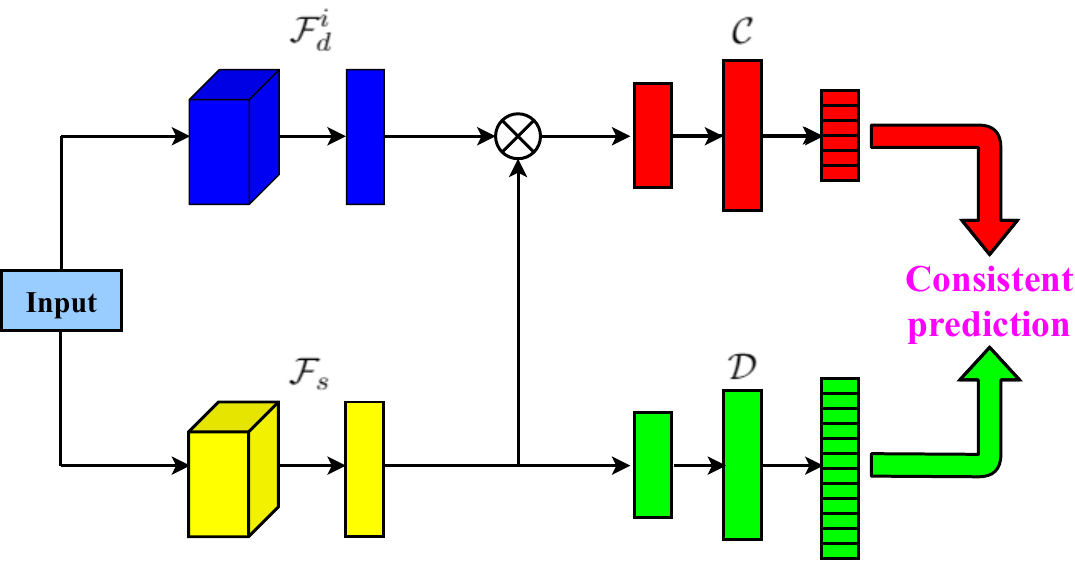}
    \caption{The architecture of the RCA model.
	}
    \label{Fig2}
\end{figure}

%%%%%%%%%%%%%%%%%%%%%%%%%%%

\subsection{Regularized Conditional Alignment}
\label{sec:RCA}

Aligning the marginal feature distributions can be accomplished using adversarial training \cite{chen2018multinomial}. The fundamental concept entails the concurrent discernment of both class and domain labels. This is achieved by training a binary classifier in tandem with an $M$-way discriminator, striving to optimize performance to the fullest extent. Simultaneously, a feature extractor endeavors to confound the discriminator. When executed effectively, the classifier adeptly distinguishes labeled data, while the discriminator attains domain-agnostic characteristics. Consequently, the classifier is poised to render accurate predictions concerning unlabeled data spanning diverse domains. However, it is imperative to underscore that prior MDTC methods \cite{chen2018multinomial, wu2020dualb} predominantly align marginal feature distributions, a practice that falls short of guaranteeing successful knowledge transfer as there may exist a misalignment risk elucidated in Section \ref{sec:intro}.

The crux of our approach revolves around the imposition of not a conventional $M$-way adversarial loss for domain alignment, but rather a $2M$-way adversarial loss. This perspective operates under the assumption of the existence of $2M$ potential domain labels, with the first $M$ denoting domains characterized by positive sentiment and the latter $M$ signifying domains imbued with negative sentiment. We name the $2M$-way discriminator as the joint discriminator, tasked with assimilating distributions across domains while accommodating diverse sentiments. The output of the joint discriminator serves a dual purpose, capable of both designating the domain and sentiment labels of the input data.

During training, given the abundance of unlabeled data spanning diverse domains, we enlist the binary classifier to furnish pseudo-labels for the unlabeled data, thereby facilitating alignment via the joint discriminator. The classifier undergoes training on all labeled data using classification loss and unlabeled data through SSL regularizers.

As illustrated in Figure \ref{Fig2}, the RCA model consists of four components: a shared feature extractor $\mathcal{F}_s$, $M$ domain-specific feature extractors $\{\mathcal{F}_d^i\}_{i=1}^M$, a classifier $\mathcal{C}$ and a joint discriminator $\mathcal{D}$. The shared feature extractor $\mathcal{F}_s$ aims to learn domain-invariant features, while the domain-specific feature extractor $\mathcal{F}_d^i$ is tailored to capture domain-specific knowledge of the $i$-th domain. The classifier $\mathcal{C}$ takes the concatenation of the shared feature and domain-specific feature and outputs the sentiment probability of an input $\xvec^i$ as:
\begin{align}
    f_c(\xvec^i)=\mathcal{C}([\mathcal{F}_s(\xvec^i),\mathcal{F}_d^i(\xvec^i)])\in \mathbb{R}^2
\end{align}

\noindent where $[\cdot,\cdot]$ represents the concatenation of two vectors. The joint discriminator $\mathcal{D}$ takes the shared feature of $\xvec^i$ as input and its output can be written as:

\begin{align}
    f_d(\xvec^i)=\mathcal{D}(\mathcal{F}_s(\xvec^i))\in\mathbb{R}^{2M}
\end{align}

\subsection{Loss Functions}

In RCA, the classifier $\mathcal{C}$ is used for evaluation by training on labeled data across different domains using the cross-entropy loss. The classification loss can be defined as:

\begin{align}
    \mathcal{L}_c=\sum_{i=1}^M\mathbb{E}_{(\xvec^i,y^i)\sim\mathbb{L}_i}\ell_{CE}(f_c(\xvec^i),y^i)
\end{align}

\noindent where $\ell_{CE}(\cdot,\cdot)$ indicates the cross-entropy loss.

The joint discriminator is trained with both labeled and unlabeled data among the $M$ domains. The $2M$-way adversarial loss is defined as:

\begin{align}
    \mathcal{L}_d=\sum_{i=1}^M\mathbb{E}_{\xvec^i\sim\mathbb{L}_i\cup\mathbb{U}_i}\ell_{CE}(f_d(\xvec^i),\tilde{d}^i)
\end{align}

\noindent For a labeled input $\xvec^i$, in the case of a positive label, its corresponding $\tilde{d}^i$ is structured as $[d^i,\mathbf{0}]$, where $d^i$ signifies the domain label of $\xvec^i$, and $\mathbf{0}$ denotes the zero vector with dimensions of $M$. Specifically, for positively labeled inputs, we designate the final $M$ joint probabilities within $\tilde{d}^i$ as zero. In the case where $\xvec^i$ is negative, its associated $\tilde{d}^i$ is configured as $[\mathbf{0},d^i]$, attributing the domain label $d^i$ to the negative input while assigning zeros to the preceding $M$ joint probabilities. Unlabeled inputs, on the other hand, have their $\tilde{d}^i$ determined by their pseudo-label. The formulation of $\tilde{d}^i$ for unlabeled data mirrors that of labeled data. It is crucial to emphasize that $\mathcal{L}_d$ serves a dual purpose. It not only guides adversarial alignment but also enforces consistent label predictions, ensuring concordance between the classifier and the joint discriminator.

\subsection{Semi-Supervised Learning Regularizers}
\label{sec:SSL}

MDTC presents itself as a dual-pronged challenge. The initial endeavor involves the mitigation of disparities that manifest among diverse domains. Following the successful alignment of these domains, the MDTC transitions into an SSL problem. Nonetheless, the conundrum of scarce labeled data resources often emerges as a bottleneck, impeding the optimal training of both a proficient classifier and a discriminator. Moreover, as we perform joint adversarial alignment involving domains and classes in RCA, the quality of pseudo-labels emerges as a pivotal concern. Subpar pseudo-labels have the potential to erode the alignment process, consequently deteriorating the discriminative power of the model.

To alleviate the above adverse effects, we employ entropy minimization \cite{grandvalet2005semi} and virtual adversarial training (VAT) \cite{miyato2018virtual}. Entropy minimization adeptly governs prediction uncertainty pertaining to unlabeled data within SSL \cite{grandvalet2005semi,berthelot2019mixmatch}, it effectively coerces decision boundaries to reside within the sparser regions of the latent space, a highly coveted attribute consonant with the cluster assumption \cite{grandvalet2005semi}. The entropy minimization loss is defined as:

\begin{align}
    \mathcal{L}_{e}=-\sum_{i=1}^M\mathbb{E}_{\xvec^i\sim\mathbb{U}_i}[f_c(\xvec^i)^{\top}\log(f_c(\xvec^i))]
\end{align}

\noindent  Since the joint discriminator undergoes training on both labeled and unlabeled data, optimizing the joint alignment of domains and classes can be achieved by processing low-entropy predictions generated by the classifier \cite{long2018conditional}. Furthermore, entropy minimization satisfies the cluster assumption only for the Lipschitz classifiers \cite{grandvalet2005semi}. To fulfill this requirement, we incorporate VAT into our model. VAT promotes the refinement of label predictions by instilling robustness in the classifier against local adversarial input perturbations $\epsilon$ \cite{miyato2018virtual}. The VAT loss on unlabeled data is defined as:

\begin{align}
    \mathcal{L}_{uvt}=\sum_{i=1}^M\mathbb{E}_{\xvec^i\sim\mathbb{U}_i}[\max_{\|r\|\leq\epsilon}D_{kl}(f_c(\xvec^i)\|f_c(\xvec^i+r))]
\end{align}

\noindent where $D_{kl}(\cdot\|\cdot)$ is the Kullback-Leibler divergence \cite{van2014renyi}. Following \cite{shu2018dirt}, we also apply VAT on labeled data:

\begin{align}
    \mathcal{L}_{lvt}=\sum_{i=1}^M\mathbb{E}_{\xvec^i\sim\mathbb{L}_i}[\max_{\|r\|\leq\epsilon}D_{kl}(f_c(\xvec^i)\|f_c(\xvec^i+r))]
\end{align}

\subsection{Overall Objective Function}
\label{sec:Fun}

We combine the introduced objective functions to define the overall objective function of the RCA method:

\begin{equation}
    \begin{aligned}
    \min_{\mathcal{F}_s,\{F_d^i\}_{i=1}^M,\mathcal{C}}\max_{\mathcal{D}}\mathcal{L}_c&+\lambda_d\mathcal{L}_d+\lambda_{uvt}(\mathcal{L}_e+\mathcal{L}_{uvt}) \\
    &+\lambda_{lvt}\mathcal{L}_{lvt}
    \end{aligned}
\end{equation}

\noindent where $\lambda_d$, , $\lambda_{uvt}$, and $\lambda_{lvt}$ are hyperparameters that trader-off different objective functions. The RCA method optimizes the model parameters in an alternating fashion following \cite{goodfellow2014generative}.

\section{Experiments}
\label{sec:Exp}

\subsection{Datasets}
\label{sec:data}

In our experiments, we conduct experiments on two MDTC benchmarks: the Amazon review dataset \cite{blitzer2007biographies} and the FDU-MTL dataset \cite{liu2017adversarial}. The Amazon review dataset encompasses four domains: books, DVDs, electronics, and kitchen. Each domain contains 1,000 positive samples and 1,000 negative samples. This dataset has undergone meticulous preprocessing, transforming textual data into a bag of features, which regrettably strips away the inherent word order information. Thus, our approach involves taking the 5,000 most prevalent features and representing each review as a 5,000-dimensional feature vector, wherein feature values are the raw counts of the features. The FDU-MTL dataset, on the other hand, presents a more formidable challenge. It spans 16 domains: 14 Amazon review domains (books, electronics, DVDs, kitchen, apparel, camera, health, music, toys, video, baby, magazine, software, and sport) and two movie review domains (IMDB and MR). The textual content within this dataset remains in its pristine form, tokenized by the Stanford tokenizer \cite{manning2014stanford}. Each domain is endowed with a development set comprising 200 samples and a test set featuring 400 samples. The number of training and unlabeled data exhibits slight fluctuations across domains, they typically hover around 1,400 and 2,000, respectively.

\subsection{Implementation Details}
\label{sec: Imp}

All experiments are executed employing PyTorch. To ensure fair comparisons, we adhere rigorously to the standard MDTC evaluation protocols \cite{chen2018multinomial}. We have three hyperparameters: $\lambda_d$, $\lambda_{uvt}$, and $\lambda_{lvt}$. In our experiments, we set $\lambda_d=0.5$, $\lambda_{uvt}=1$, and $\lambda_{lvt}=0.01$. The batch size is set to $8$, and training is orchestrated using the Adam optimizer with a learning rate of $0.0001$. ReLU serves as the activation function, the number of training iterations is 50, and the dropout rate for each component is 0.4. Furthermore, we adopt the same network architecture following \cite{chen2018multinomial,wu2021conditional}. We set the dimension of the domain-invariant feature as 128 and that of the domain-specific feature as 64. The classifier $\mathcal{C}$ is a multi-layer perceptron (MLP) with one hidden layer containing $128+64$ units. The joint discriminator is also an MLP with one hidden layer whose size is $128$.

For the Amazon review dataset, we employ MLPs featuring two hidden layers with 1,000 and 500 units, respectively, as feature extractors. The input dimension stands at 5,000. We conduct 5-fold cross-validation on this dataset and report the average test classification accuracy across the five folds.

For the FDU-MTL dataset, we employ a Convolutional Neural Network (CNN) with one convolutional layer to serve as a feature extractor. The CNN is designed to encompass different kernel sizes $(3,4,5)$, with a total of 200 kernels. Notably, the input to the convolutional layer comprises 100-dimensional word embeddings, obtained by using word2vec \cite{mikolov2013efficient} for each word within the input sequence.

\subsection{Comparison Baselines}
\label{sec:base}

Our RCA method is compared with the following baselines. The collaborative multi-domain sentiment classification (CMSC) combines the outputs of a shared classifier and a set of domain-dependent classifiers to make the final predictions \cite{wu2015collaborative}. The CMSC models can be trained with the least square loss (CMSC-LS), the hinge loss (CMSC-SVM), and the log loss (CMSC-Log). The adversarial multi-task learning for text classification (ASP-MTL) uses adversarial training and long short-term memory (LSTM) networks to capture share-private separation of different domains \cite{liu2017adversarial}. The multinomial adversarial network (MAN) leverages two loss functions to train the discriminator: the least square loss (MAN-L2) and the negative log-likelihood loss (MAN-NLL) \cite{chen2018multinomial}. The multi-task learning with a bi-directional language model (MT-BL) employs language modeling and a uniform label distribution-based loss function to guide the feature learning \cite{yang2019multi}. The dual adversarial co-learning (DACL) combines two types of adversarial training, discriminator-based and classifier-based adversarial training \cite{wu2020dualb}. The conditional adversarial network (CAN) performs the alignment of conditional feature distributions \cite{wu2021conditional}. All compared methods follow the standard MDTC evaluation protocols, we hence conveniently cite the results from \cite{wu2020dualb,wu2021conditional}.

%%%%%

\subsection{Results}
\label{sec:res}

%%%%%%%%%%%%%%%%%%%%%%%%%%%%%%%%%%%%%%%%%%%%%%%%%

\begin{table}[t]
\caption{\label{font-table} MDTC results on the Amazon review dataset.}\smallskip
\label{table_ref1}
\centering
\resizebox{1.0\columnwidth}{!}{
\smallskip\begin{tabular}{ l|  c c c c c c c c c}
\hline
	Domain & CMSC-LS & CMSC-SVM & CMSC-Log & MAN-L2 & MAN-NLL & DACL & CAN & RCA(Proposed)\\
\hline
Books &  82.10 & 82.26 & 81.81 & 82.46 & 82.98 & 83.45 & 83.76 & $\mathbf{84.97\pm0.15}$ \\
DVD &  82.40 & 83.48 & 83.73 & 83.98 & 84.03 & 85.50 & 84.68 & $\mathbf{85.81\pm0.11}$ \\
Electr.  & 86.12 & 86.76 & 86.67 & 87.22 & 87.06 & 87.40 & 88.34 &  $\mathbf{89.35\pm0.17}$ \\
Kit.  &  87.56 & 88.20 & 88.23 & 88.53 & 88.57 & 90.00 & 90.03 & $\mathbf{90.86\pm0.09}$\\
\hline
AVG  &  84.55 & 85.18 & 85.11 & 85.55 & 85.66 & 86.59 & 86.70 & $\mathbf{87.75\pm0.12}$\\
\hline
\end{tabular}}
\end{table}
%%%%%%%%%%%%%%%%%%%%%%%%%%%%%%%%%%%%%%%%
\begin{table}[t]
\caption{\label{font-table} MDTC results on the FDU-MTL dataset.}\smallskip
\label{table_ref2}
\centering
\resizebox{1.0\columnwidth}{!}{
\begin{tabular}{ l| c c c c c c c c}
\hline
Domain & ASP-MTL & MAN-L2 & MAN-NLL & MT-BL & DACL & CAN & RCA(Proposed)\\
\hline
books & 84.0 & 87.6 & 86.8 & 89.0 & 87.5 & 87.8 & $\mathbf{89.2\pm0.4}$ \\
electronics & 86.8 & 87.4 & 88.8 & 90.2 & 90.3 & $\mathbf{91.6}$ & 87.6$\pm$0.5 \\
dvd & 85.5 & 88.1 & 88.6 & 88.0 & 89.8 & 89.5 & $\mathbf{90.4\pm0.3}$ \\
kitchen & 86.2 & 89.8 & 89.9 & 90.5 & 91.5 & 90.8 & $\mathbf{92.5\pm0.4}$\\
apparel & 87.0 & 87.6 & 87.6 & 87.2 & $\mathbf{89.5}$ & 87.0 & 88.6$\pm$0.5\\
camera & 89.2 & 91.4 & 90.7 & 89.5 & 91.5 & 93.5 & $\mathbf{95.8\pm0.2}$\\
health & 88.2 & 89.8 & 89.4 & $\mathbf{92.5}$ & 90.5 & 90.4 & 90.7$\pm$0.4 \\
music & 82.5 & 85.9 & 85.5 & 86.0 & 86.3 & 86.9 & $\mathbf{87.7\pm0.3}$ \\
toys & 88.0 & 90.0 & 90.4 & 92.0 & 91.3 & 90.0 & $\mathbf{92.4\pm0.3}$ \\
video & 84.5 & 89.5 & 89.6 & 88.0 & 88.5 & 88.8 & $\mathbf{90.0\pm0.2}$\\
baby & 88.2 & 90.0 & 90.2 & 88.7 & 92.0 & 92.0 & $\mathbf{92.7\pm0.6}$ \\
magazine & 92.2 & 92.5 & 92.9 & 92.5 & 93.8 & 94.5 & $\mathbf{95.4\pm0.4}$ \\ 
software & 87.2 & 90.4 & 90.9 & 91.7 & 90.5 & 90.9 & $\mathbf{92.9\pm0.5}$ \\
sports & 85.7 & 89.0 & 89.0 & 89.5 & 89.3 & $\mathbf{91.2}$ & 89.7$\pm$0.5 \\
IMDb & 85.5 & 86.6 & 87.0 & 88.0 & 87.3 & 88.5 & $\mathbf{89.5\pm0.1}$\\
MR & 76.7 & 76.1 & 76.7 & 75.7 & 76.0 & 77.1 & $\mathbf{78.3\pm0.3}$\\
\hline
AVG & 86.1 & 88.2 & 88.4  & 88.6 & 89.1 & 89.4 & $\mathbf{90.2\pm0.3}$ \\
\hline
\end{tabular} 
}
\end{table}
%%%%%%%%%%%%%%%%%%%%%%%%%

In this paper, we present experimental results based on 5 runs. The experimental results on the Amazon review dataset are showcased in Table \ref{table_ref1}. Notably, our RCA method attains the highest average accuracy of $87.75\%$, surpassing CAN by a significant margin of $1.05\%$. In addition, our model excels in every individual domain, eclipsing the performance of competing baselines.

The experimental results of the FDU-MTL dataset are presented in Table \ref{table_ref2}. From Table \ref{table_ref2}, it can be noted that our RCA model obtains the best average accuracy, beating the runner-up CAN model by a margin of 0.8\%. Impressively, our method outperforms other baselines on 12 out of 16 domains.

\section{Conclusion}
\label{sec:con}

In this paper, we propose a Regularized Conditional Alignment (RCA) method for MDTC. Our RCA method distinguishes itself by aligning the joint distributions of domains and classes through a sophisticated enhancement of the conventional adversarial loss, leading to a refined $2M$-way adversarial loss. This refined joint alignment serves as an effective safeguard against the misalignment risk brought by aligning marginal feature distributions in MDTC. Furthermore, we enrich our approach by integrating entropy minimization and virtual adversarial training to facilitate the displacement of decision boundaries from the densely populated regions in the shared latent space and impose the Lipschitz constraint on our model, respectively. The experimental results on two MDTC benchmarks unequivocally demonstrate that our RCA method can outperform the state-of-the-art baselines.

\bibliographystyle{IEEEbib}
\bibliography{strings}

\end{document}